\pdfoutput=1
\documentclass[11pt]{article}

\usepackage{ACL2023}

\usepackage{times}
\usepackage{latexsym}
\usepackage{graphicx}
\usepackage{color, colortbl}
\usepackage{multirow}
\usepackage[T1]{fontenc}

\usepackage[utf8]{inputenc}

\usepackage{microtype}

\usepackage{inconsolata}

\usepackage{comment}

\title{First Attempt at Building Parallel Corpora for Machine Translation of Northeast India's Very Low-Resource Languages}

\author{
\normalsize Atnafu Lambebo Tonja$^{1,2,3}$, Melkamu Mersha $^{1}$, Ananya Kalita$^{4}$, \\
\textbf{\normalsize Olga Kolesnikova $^{2}$ and Jugal Kalita $^{1}$  } \\\\
 \footnotesize
 $^{1}$ University of Colorado, Colorado Springs, USA
$^{2}$ Instituto Politécnico Nacional, Mexico city, Mexico, \\
 \footnotesize
$^{3}$ Lelapa AI, South Africa,
$^{4}$ Palmer Ridge High School, Monument, Colorado, USA
 }

\begin{document}
\maketitle
\begin{abstract}
This paper presents the creation of initial bilingual corpora for thirteen very low-resource languages of India, all from Northeast India. 
It also presents the results of initial translation efforts in these languages. It creates the first-ever parallel corpora for these languages and provides initial benchmark neural machine translation results for these languages.
We intend to extend these corpora to include a large number of low-resource Indian languages and integrate the effort with our prior work with African and American-Indian languages to create corpora covering a large number of languages from across the world. 
\end{abstract}

 \begin{figure*} 
 \small
\centering
\includegraphics[width=6in,height=3.9in]{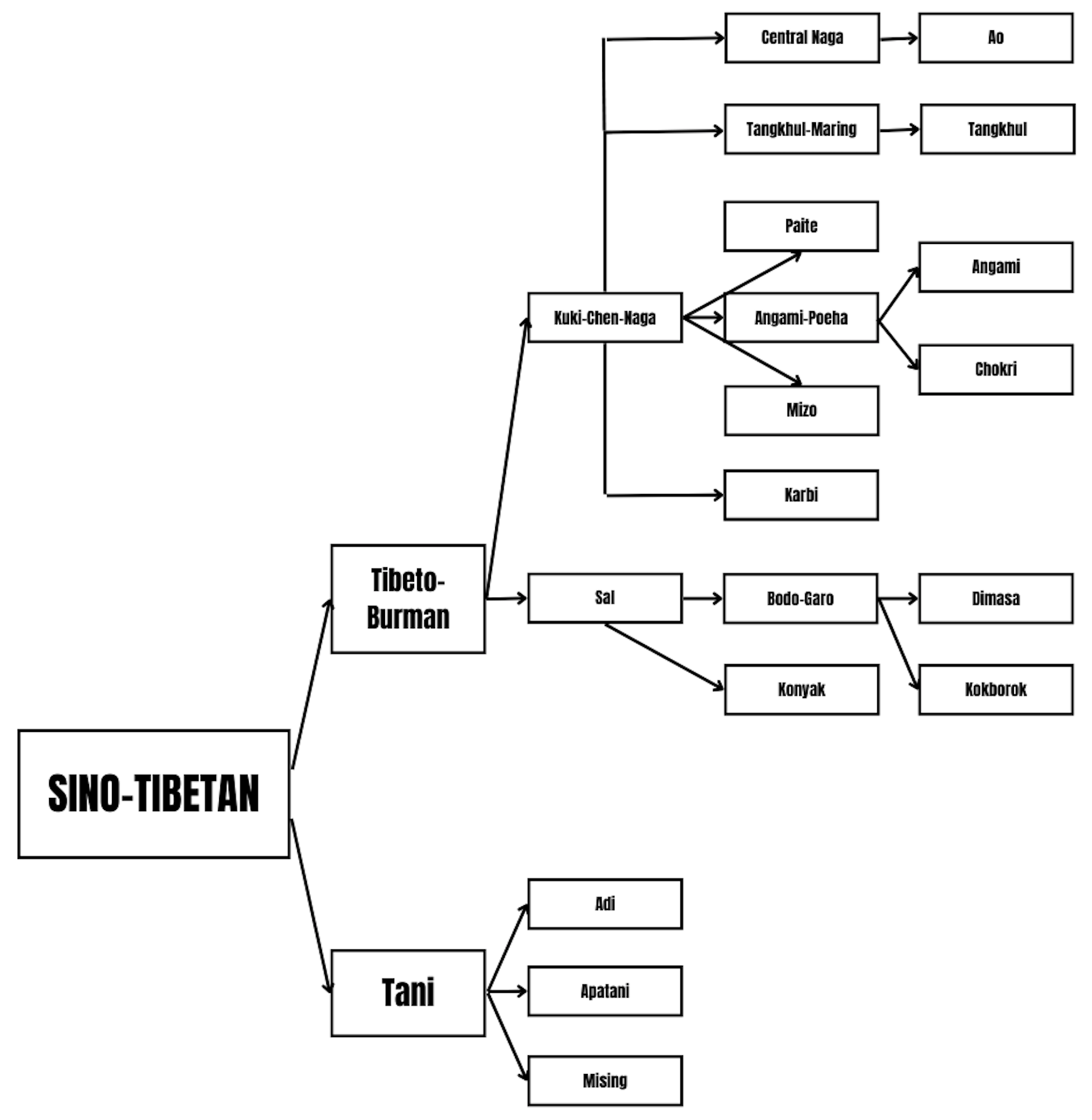}
\caption{The languages under consideration fall under the Sino-Tibetan family and two subfamilies, Tani and Tibeto-Burman.  }
\label{fig:languageTree}
\end{figure*}

\section{Introduction}
In the last few years, there have been significant advancements in deep learning approaches, such as the development of transformers \cite{vaswani2017attention}. These advancements have greatly improved machine translation (MT), a core natural language processing (NLP) task. Regarding coverage and translation quality, MT has shown remarkable improvements \cite{wang2021progress}. Most models and methods for high-resource languages do not perform well in low-resource settings. Low-resource languages have also suffered from inadequate language technology designs \cite{costa2022no,tonja2023natural,king2015practical,joshi2019unsung,tonja2022improving,tonja2023low,yigezu2021multilingual}. Creating effective methods for natural language tasks is challenging, with extremely limited resources and little to no available data. The problem becomes worse without a parallel dataset for a vast number of world's languages \cite{joshi2020state, ranathunga2023neural, adebara-abdul-mageed-2022-towards}.

Ethnologue\footnote{www.ethnologue.com} enumerates 7,618 extant human languages in the world, of which it estimates 3,072 are endangered. 
India has 30 languages spoken by more than a million native speakers, 92 additional languages spoken by more than 10,000 people, and 1599 other languages. The terms language and dialect may have been conflated in Census of India reports\footnote{https://en.wikipedia.org/wiki/Languages\_of\_India}.
Of these, 197  languages are endangered \cite{dash-etal-2022-permutation}. The languages with which work in this paper are a small fraction of such languages. 

This paper makes a selection of low-resource languages of India, all from Northeast India, with the intention of developing resources to facilitate neural machine translation. 
We choose Northeast India because this area has a very high concentration of very low-resource and/or endangered languages, and the area being remote from mainland India has exacerbated scholarly neglect. 

\section{Related Work}
\citet{dash2020revitalizing} reviewed the issues languages with a small number of speakers face in India. 
Based on extensive research, the author surmised that the amount of published work on low-resource and endangered languages of India, especially in the context of technology and, in particular, natural language processing, is miniscule. 

\citet{chauhan2021monolingual}  provided a monolingual corpus of Kangri (a language spoken by 1.1 million people in Himachal Pradesh and Punjab) with 1.8M words, as well bitext Hindi-Kangri corpus of 27 thousand words.  
For NMT, they used a model that learns to map word embedding from one language to another in an unsupervised manner \cite{artetxe2018robust}. 
Only cursory details are given regarding the architecture of the neural model, although translation metrics are given in terms of BLEU and Meteor scores. The corpora are available on Github\footnote{https://github.com/chauhanshweta/Kangri\_corpus}.

\citet{acharya-2021-kaushikacharya} discussed a proposal for 
building a digital archive to collect and preserve textual, audio, and video documentation of twelve Mund\-{a} languages, a sub-family of Austro-Asiatic languages. Of twenty Mund\-{a} languages, spoken by between just 20 to 7M people, UNESCO has identified twelve as endangered.  The authors proposed to use advanced technologies like artificial intelligence in the design of the archive. However, it is not clear what has been achieved. 

To  the authors' knowledge, there is no prior computational work of any kinda in the very low-resource and endangered languages of Northeast India.

\section{Languages Under Consideration}

The Sino-Tibetan language family includes more than 400 modern languages spoken in China, India, Burma, and Nepal. It is one of the most diverse language families in the world, with 1.4 billion speakers, including Chinese, Tibetan, and Burmese.
 Based on a phylogenetic study of 50 ancient and modern Sino-Tibetan languages,  scholars have recently concluded that the Sino-Tibetan languages originated in North China around 7,200 years ago \cite{sagart2019dated}. 
 Various classification schemes have been proposed for the Sino-Tibetan family of languages \cite{matisoff2003handbook,matisoff2015sino,driem2001languages}. 
 Our discussion follows Matisoff's classification that divides Sino-Tibetan languages into a number of sub-families at various levels. 

 \begin{table*}
 \small
     \centering
     \begin{tabular}{l|l|p{1in}|r|p{1in}|p{0.5in}|p{0.5in}} \hline 
Language & ISO Code & Family &  Speakers & Location & Corpus Domain &Corpus Size\\ \hline 
 Adi &adi &Tani&  150K&Arunachal Pradesh &Religious&29301\\\hline
 Angami & njm&Tibeto-Burman&  150K&Nagaland & " &30017\\\hline
 Ao & njo&Tibeto-Burman&  607K &Nagaland&  " &29121\\ \hline 
 Apatani & apt&Tani&  45K&Arunachal Pradesh&"&7185 \\\hline
Chokri & nri& Tibeto-Burman &  111K&Nagaland&  "&7821\\\hline
Dimasa & dis&Tibeto-Burman&  137K&Assam, Nagaland &"&10275\\\hline
Karbi & mjw&Tibeto-Burman&  2.5M &Assam, Meghalaya, Arunachai Pradesh&   "&7185\\ \hline 
Kokborok & trp&Tibeto-Burman&  1M &Tripura, Assam, Mizoram, Myanmar, Bangladesh&   "&29298\\ \hline 
Konyak & nbe&Tibeto-Burman&  246K& Nagaland, Myanmar&" &28518\\\hline
Mising & mrg&Tani&  629K &Assam& " &7825\\ \hline 
Paite & pck&Tibeto-Burman&  1M &Manipur, Mizoram, Assam, Myanmar&   "&29615\\ \hline 
 Tangkhul & nmf&Tibeto-Burman& 140K &Manipur, Nagaland& "&28324\\\hline
Thado & tcz&Tibeto-Burman&  350K &Manipur, Nagaland, Assam, Mizoram&  "&29004\\ \hline 


 
     \end{tabular}
     \caption{Languages Under Consideration. Accurate population count is difficult to obtain and varies substantially among sources, corpus size shows the number of parallel sentences.}
     \label{table:languages}
 \end{table*}

 Table \ref{table:languages} provides a list of languages we work with in this paper. No detailed description of the languages is given in this paper due to lack of space.  
The languages are from Northeast India, which is linguistically very diverse and has a large number of very low-resource languages that are vulnerable or endangered. 
The languages we have chosen all belong to the Sino-Tibetan family. 
Under this class, there are two sub-families called Tibeto-Burman and Taani, among other sub-families. All our languages come from these two sub-families (see Figure \ref{fig:languageTree}). 

\section{Dataset}
\subsection{Dataset Collection}

We obtained datasets in 13 Indian languages from religious domains through a Bible-related website\footnote{https://www.bible.com/}. 
To extract the Bible data from these websites, we utilized a web crawler that identified the structure of web documents, including pages, books, and phrases, for each article. Python libraries like {\em requests}, regular expressions ({\em R}), and Beautiful Soup ({\em BS}) were used to extract article content and analyze website structure from a given URL.
Extensive research did not discover any additional publicly available texts in these languages. 

\subsection{Sentence Alignment}
We gathered the limited corpora for various languages and aligned each Indian language sentence with a corresponding sentence in English to create a dataset for the MT experiment. We followed the heuristic alignment method outlined by the \citet{tonja2023parallel} to align the sentences.

\subsection{Dataset Pre-processing}
We aligned the texts of Indian languages with their corresponding translations in English. Before splitting the corpus, we pre-processed by removing numbers, special characters, and sentences that contain less than five words. We divided the pre-processed corpus into training, development, and test sets in a 70:10:20 ratio for the baseline experiments. Detailed information on the selected languages, language families, domain, and dataset size can be found in Table \ref{table:languages}.
\begin{figure}
    \centering
    \includegraphics[width=0.5\textwidth]{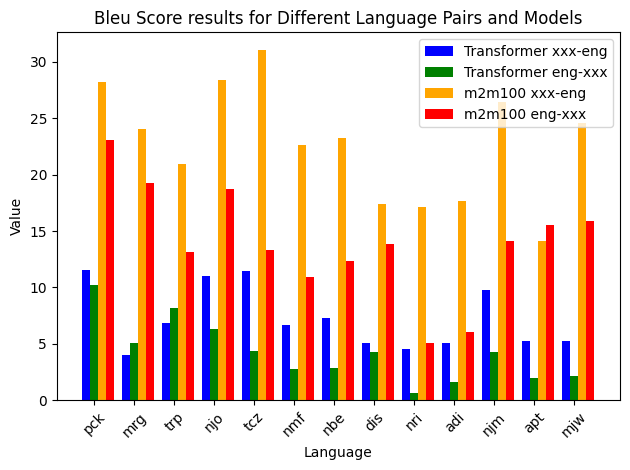}
    \caption{Benchmark translation results for transformer and fine-tuned approaches in both directions (English/low-resource Indian languages).}
    \label{fig:fig_lang}
\end{figure}

\begin{table*}[!ht]
\centering
  \resizebox{\textwidth}{!}{
\begin{tabular}{c|lllllllllllll|l}\hline
\multirow{2}{*}{Model} & \multicolumn{13}{c}{en-xx} \\
                       & \textbf{adi} & \textbf{apt} & \textbf{dis} & \textbf{mjw} & 
                       \textbf{mrg} & \textbf{nbe} & 
                       \textbf{njm} & \textbf{njo} & 
                       \textbf{nmf} & \textbf{nri} & 
                       \textbf{pck} & \textbf{tcz} & 
                       \textbf{trp} &\textbf{Avg.}\\\hline
                       & \multicolumn{13}{c}{\textbf{Bleu Score}}                                                \\\hline
Transformer  &1.62&2&4.33&2.17&5.12&2.91&4.26&6.32&2.81&0.66&10.23&4.35&9.18&4.30  \\\hline
m2m100-fine-tuned  &6.06&15.49&13.82&15.91&19.23&12.31&14.07&18.76&10.88&5.10&23.03&13.28&13.16&\textbf{13.93}  \\ \hline
\end{tabular}
}
\caption{Benchmark translation results from English to  low-resource Indian languages}
\label{tab:from_eng}
\end{table*}

\begin{table*}[!ht]
\centering
  \resizebox{\textwidth}{!}{
\begin{tabular}{c|lllllllllllll|l}\hline
\multirow{2}{*}{Model} & \multicolumn{13}{c}{xx-en} \\
                        & \textbf{adi} & \textbf{apt} & \textbf{dis} & \textbf{mjw} & 
                       \textbf{mrg} & \textbf{nbe} & 
                       \textbf{njm} & \textbf{njo} & 
                       \textbf{nmf} & \textbf{nri} & 
                       \textbf{pck} & \textbf{tcz} & 
                       \textbf{trp} &\textbf{Avg.}\\\hline
                       & \multicolumn{13}{c}{\textbf{Bleu Score}}                                                     \\\hline
Transformer   &5.09&5.29&5.11&5.3&4.02&7.32&9.8&11.06&6.64&4.51&11.54&11.49&6.88& 7.63\\\hline
m2m100-fine-tuned   &17.62&14.11&17.42&24.52&24.03&23.27&26.44&28.37&22.61&17.12&28.16&31.03&20.90&\textbf{22.67}   \\ \hline
\end{tabular}
}
\caption{Benchmark translation results from  low-resource Indian Languages to English language}
\label{tab:to_eng}
\end{table*}
\section{Baseline Experiments and Results}
\subsection{Experiments}
To evaluate the usability of the newly collected corpus, we trained bi-directional MT models that can translate Indian languages to/from English using (1) \textbf{transformer} and (2) \textbf{fine-tuning} multilingual machine translation model. 

\textbf{1) Transformer} - is a type of neural network architecture first introduced in the paper \textit{Attention Is All You Need} \cite{vaswani2017attention}. 
Transformers are state-of-the-art approaches widely used in NLP tasks such as MT, text summarization, and sentiment analysis.  We trained transformers from scratch for this experiment. 

\textbf{2) Fine-tuning} involves using a pre-trained MT model and adapting it to a specific translation task, such as translating between a particular language pair or in a specific domain \cite{lakew2019adapting}. We used \textbf{M2M100-48} a multilingual encoder-decoder (seq-to-seq) model trained for many-to-many multilingual translation \cite{fan2020beyond}. We used a model with 48M parameters due to computing resource limitations.
\subsection{Results}
We used Sacrebleu \cite{post-2018-call} evaluation metrics to evaluate translation models.
Tables \ref{tab:from_eng}, \ref{tab:to_eng} and Figure \ref{fig:fig_lang} show the translation results in both directions (to/from English - from/to  low-resource Indian languages)
\subsubsection{Translating from English to low-resource Indian languages }
In Table \ref{tab:from_eng}, we present the translation results from English to low-resource Indian languages. We observed that fine-tuning the m2m100 model performs better than using a transformer trained from scratch. The transformer model's performance also varies significantly (0.66 -- 10.23 spBLEU) depending on the language and corpus size. This indicates that a bilingual translation model trained from scratch performs poorly for low-resource language training compared to fine-tuning multilingual translation models. Fine-tuning the multilingual model produced better results than the model built from scratch for English to Indian language translation. 

\subsubsection{Translating from low-resource Indian languages to English }
Table \ref{tab:to_eng} displays the results of using English as the target language to translate  low-resource Indian languages. As is evident from the results, the fine-tuned model outperforms the transformer model significantly when translating from Indian languages to English. However, when it comes to translating similar languages to English, the transformer model shows an improvement compared to Table \ref{tab:from_eng}. It is worth noting that the fine-tuned model exhibits better Bleu scores while translating to English than when translating to low-resource Indian languages. The results indicate that languages with larger datasets tend to perform better. Therefore, both models exhibit improved performance while translating from  low-resource Indian languages to English, whereas the model struggles to translate from English to  low-resource Indian languages.

\section{Conclusions and Future Work}
This paper presents the first, albeit limited size,  parallel corpus for 13 low-resource Sino-Tibetan Indian languages paired with English and discusses the benchmark results for the translation of language pairs to/from low-resource Indian languages from/to English. We evaluated the usability of the collected corpus by using transformer and fine-tuning multilingual translation model. From our results fine-tuning multilingual model outperforms transformer model trained from scratch in both translation directions.

In the future, we aim to increase corpus sizes of  these low-resource languages by extracting text from scanned documents if/where available  and evaluate additional machine translation approaches to improve performance. 
We intend to increase the number of languages substantially by first incorporating all  low-resource languages of India for which a Bible translation exists. 
We also plan to find language communities in social media platforms such as Facebook and attempt to gather additional bitext documents and evaluate the quality of translations with native speakers. 




\bibliography{anthology,custom}
\bibliographystyle{acl_natbib}

\appendix



\end{document}